\documentclass[conference]{IEEEtran}
\IEEEoverridecommandlockouts

\newcommand{\hide}[1]{}

\usepackage[ruled,vlined,boxed,linesnumbered]{algorithm2e}
\usepackage{graphics}
\usepackage{booktabs}
\usepackage{cite}
\usepackage{amsmath,amssymb,amsfonts}
\usepackage{algorithmic}
\usepackage{graphicx}
\usepackage{textcomp}
\usepackage{xcolor}
\usepackage{multirow}
\usepackage{enumitem}
\usepackage{hyperref}
\usepackage{bm}

\newcommand{\etal}{\textit{et al}.}

\begin{document}
\title{AutoHEnsGNN: Winning Solution to AutoGraph Challenge for KDD Cup 2020}

\author{\IEEEauthorblockN{1\textsuperscript{st} Jin Xu \IEEEauthorrefmark{1}}
\IEEEauthorblockA{
\textit{Tsinghua University}\\
Beijing, China \\
j-xu18@mails.tsinghua.edu.cn}
\and
\IEEEauthorblockN{2\textsuperscript{nd} Mingjian Chen \IEEEauthorrefmark{1}}
\IEEEauthorblockA{
\textit{Meituan} \\
Beijing, China \\
chenmingjian@meituan.com}\thanks{\IEEEauthorrefmark{1}The first two authors contributed equally.}
\and
\IEEEauthorblockN{3\textsuperscript{rd} Jianqiang Huang}
\IEEEauthorblockA{
\textit{Meituan} \\
Beijing, China \\
huangjianqiang@meituan.com}
\and
\IEEEauthorblockN{4\textsuperscript{th} Xingyuan Tang}
\IEEEauthorblockA{
\textit{Meituan} \\
Beijing, China \\
tangxingyuan@meituan.com}
\and
\IEEEauthorblockN{5\textsuperscript{th} Ke Hu}
\IEEEauthorblockA{
\textit{Meituan} \\
Beijing, China \\
huke05@meituan.com}
\and
\IEEEauthorblockN{6\textsuperscript{th} Jian Li}
\IEEEauthorblockA{
\textit{Tsinghua University}\\
Beijing, China \\
lijian83@mail.tsinghua.edu.cn}
\and
\IEEEauthorblockN{7\textsuperscript{th} Jia Cheng}
\IEEEauthorblockA{
\textit{Meituan} \\
Beijing, China \\
jia.cheng.sh@meituan.com}
\and
\IEEEauthorblockN{8\textsuperscript{th} Jun Lei}
\IEEEauthorblockA{
\textit{Meituan} \\
Beijing, China \\
leijun@meituan.com}
}

\maketitle

\begin{abstract}
Graph Neural Networks (GNNs) have become increasingly popular and achieved impressive results in many graph-based applications. However, extensive manual work and domain knowledge are required to design effective architectures, and the results of GNN models have high variance with different training setups, which limits the application of existing GNN models. 
In this paper, we present AutoHEnsGNN, a framework to build effective and robust models for graph tasks without any human intervention. AutoHEnsGNN won first place in the AutoGraph Challenge for KDD Cup 2020, and achieved the best rank score of five real-life datasets in the final phase. 
Given a task, AutoHEnsGNN first applies a fast proxy evaluation to automatically select a pool of promising GNN models. Then it builds a hierarchical ensemble framework: 1) We propose graph self-ensemble (GSE), which can reduce the variance of weight initialization and efficiently exploit the information of local and global neighborhoods; 2) Based on GSE, a weighted ensemble of different types of GNN models is used to effectively learn more discriminative node representations.
To efficiently search the architectures and ensemble weights, we propose AutoHEnsGNN$_{\text{Gradient}}$, which treats the architectures and ensemble weights as architecture parameters and uses gradient-based architecture search to obtain optimal configurations, and AutoHEnsGNN$_{\text{Adaptive}}$, which can adaptively adjust the ensemble weight based on the model accuracy.
Extensive experiments on node classification, graph classification, edge prediction and KDD Cup challenge demonstrate the effectiveness and generality of AutoHEnsGNN.
\end{abstract}

\begin{IEEEkeywords}
graph neural network, automated, hierarchical ensemble, kddcup
\end{IEEEkeywords}

\section{Introduction}
Graphs have been widely applied in real-world  applications to model interacting objects, such as social networks~\cite{derr2018relevance, DBLP:conf/cikm/QiHZCL21}, social media~\cite{liu2012probabilistic}, knowledge graphs~\cite{aggarwal2010managing}, 
stocks in the financial market~\cite{xu2020adaptive} and blockchain~\cite{li2020dissecting, AbayAGKITT19}. Recently, deep learning on the graph-structured data, named Graph Neural Network (GNN), has become increasingly popular and produced impressive results in many real-life graph applications~\cite{yang2016revisiting,hu2020open, wang2020traffic}. 
Many GNN models~\cite{velivckovic2018graph,kipf2016semi,hamilton2017inductive,xu2018powerful,ying2018graph,derr2018signed,Xu_symbiosis} have been proposed and have achieved start-of-the-art results in diverse domains~\cite{ying2018graph,derr2018signed,zitnik2018modeling}.

Despite the success of advanced GNN models, there is no single winner that can dominate others across different tasks~\cite{shchur2018pitfalls,hu2020open}. To build an effective model for practical usage, two problems need to be addressed: 1) Extensive manual efforts and domain knowledge are required to design an effective architecture, which can be expensive to design architectures for numerous applications; 2) For GNN models, the nodes in the graph are not mutually independent and share the same trainable weights. Thus, the 
performance of GNN models has high variance with different initialization or training/validation splits of the graph, 
called pitfalls of GNN evaluation (see~\cite{shchur2018pitfalls}).
For real-life scenarios, we need to manually (usually randomly) decide the split of training and validation, and obtain reliable predictions for the unseen test set. Without considering variance caused by these factors, models may not generalize well to the unseen test set. 
These problems hinder real-life applications of the existing GNN models due to the demand of human experts and the variance of GNN models. Therefore, there is a strong demand for automatically designing and training GNN models which can deliver robust and accurate predictions for various tasks.

Previous work on automated GNN~\cite{gao2020graph,zhou2019auto,nunes2020neural,zhao2020simplifying, anonymous2021efficient} mainly focuses on searching a single novel architecture (e.g., aggregate function, activate function), which has achieved impressive performance on the public benchmark~\cite{yang2016revisiting}. However, reinforcement learning~\cite{gao2020graph,zhou2019auto,nunes2020neural} and evolutionary algorithms~\cite{nunes2020neural} based micro-search on a single model usually require significant computational resources, which is unaffordable for practical deployment. One-shot supernet based methods~\cite{zhou2019auto,anonymous2021efficient} mainly utilize weight sharing to improve efficiency but the weight sharing is quite sensitive to the search space~\cite{pourchot2020share,shu2019understanding,xie2020weight,yu2020train}, which still requires careful design of search space by human experts. In addition, existing work does not consider the variance of the GNN model, which is important for real-life applications.

To resolve the above issues, we propose Automated Hierarchical Ensemble of Graph Neural Network (AutoHEnsGNN) to automatically ensemble a group of effective models for graph tasks without any human intervention. 
Rather than searching novel architectures for individual models, the core idea is that ensemble of multiple models can be much more effective and robust than individual models~\cite{wang2020multiple, wang2003mining, puuronen2001correlation, puuronen2001ensemble}. Specifically, we first propose a novel proxy evaluation method to efficiently rank the existing models and select a pool of models with high performance for a given task. Our experiments demonstrate that the model selection accelerated by proxy evaluation can be 5$\sim$10 $\times$ faster than the accurate selection, and can find top-performing models with a high Kendall rank correlation coefficient. Then, we propose graph self-ensemble (GSE), which constructs the ensemble of the same type of models from the pool with different initializations to reduce the variance and different layers to exploit the information from local and global neighborhoods. Real-world graphs usually have a combination of several properties, which are difficult to capture by a single approach. Thus, we further apply a weighted ensemble of different types of models in the pool to aggregate multiple graph representation. To efficiently search the layers and ensemble weights, we propose a gradient-based algorithm called AutoHEnsGNN$_{\text{Gradient}}$ by treating them as architecture parameters and formulating the search as a bi-level optimization problem~\cite{anandalingam1992hierarchical,colson2007overview,liu2018darts}. Then we apply time-efficient first-order approximation to iteratively update the weights of the model and architectures. Besides, we 
propose AutoHEnsGNN$_{\text{Adaptive}}$, which can adaptively adjust the ensemble weights according to the performance of models in the pool. It can avoid co-training different types of models and is tailored for scenarios with limited GPU memory. In addition, we also construct bagging of models trained on the different splits of the dataset to further reduce the variance. 

We summarize our contributions as follows:
\begin{itemize}[leftmargin=*]
    \item We propose a novel and practical framework, named AutoHEnsGNN, which can automatically ensemble a pool of effective models to enable robust and accurate predictions for graph tasks without any human intervention.
    \item To improve the efficiency, we propose a proxy evaluation method for automatic model selection, and a gradient-based optimization algorithm for searching ensemble weights and model configurations. Besides, for scenarios with limited GPU memory, we propose a method to adaptively adjust the ensemble weights.
    \item In the AutoGraph challenge for KDD Cup 2020, AutoHEnsGNN won 1st place and achieved the best rank score of five datasets on average in the final phase. The ablation study and extensive experiments also show the effectiveness of AutoHEnsGNN.
\end{itemize}

\section{Related Work}
\begin{figure*}[!t]
\vspace{-1mm}
\centering
\includegraphics[width=0.95\textwidth]{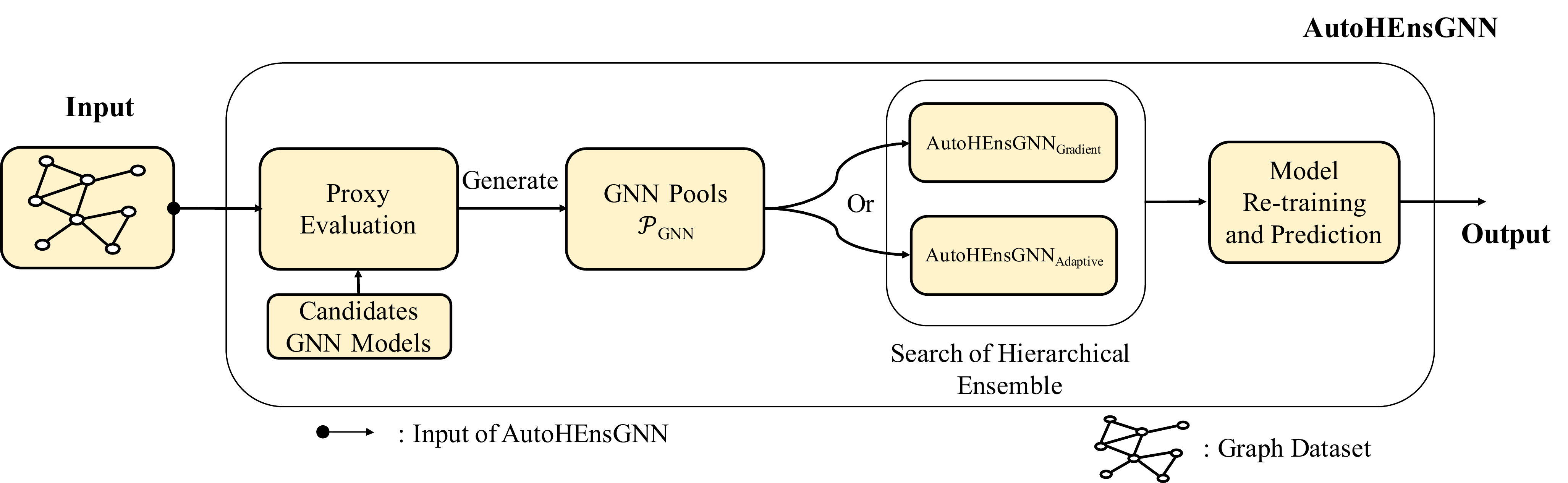}
\vspace{-1mm}
\caption{Overview of AutoHEnsGNN. Given an input graph dataset, AutoHENsGNN automatically builds models and makes predictions without any human intervention. First, a proxy evaluation method is conducted to efficiently select several top-performing GNN models $\mathcal{P}_{\text{GNN}}$ from numerous candidate ones. Then, these top-performing models can be organized in a way of the hierarchical ensemble. Two algorithms AutoHEnsGNN$_{\text{Adaptive}}$ and AutoHEnsGNN$_{\text{Gradient}}$ can be used to search for the hierarchical ensemble's configurations such as ensemble weights and architecture parameters. Finally, top-performing models with searched optimal configurations from scratch are re-trained  and aggregated in the way of the hierarchical ensemble for effective and robust predictions.}
\label{fig:overview}
\vspace{-1mm}
\end{figure*}

\subsection{Graph Neural Network}
Graph Neural Network~(GNN)~\cite{hamilton2017inductive,morris2019weisfeiler,liu2020decoupled,wang2020global,wu2019simplifying,zeng2021graph,chiang2019cluster}, as a new type of neural network for learning over graphs, has shown its superiority over graph-agnostic neural networks on diverse tasks~\cite{kipf2016semi,xu2018powerful,hamilton2017inductive,dwivedi2020benchmarking}.
Many designs of aggregators such as convolutional aggregator~\cite{kipf2016semi,defferrard2016convolutional,hamilton2017inductive}, attention aggregator~\cite{velivckovic2018graph}, gate updater~\cite{li2015gated} and skip connection~\cite{fey2019just} have been proposed and achieved impressive results in diverse domains~\cite{ying2018graph,derr2018signed,zitnik2018modeling}. 
Each type of design can exploit a specific structural property~\cite{goyal2019graph}. For example, GIN~\cite{xu2018powerful}, with injective aggregating function, can better capture the structure properties, GCNII~\cite{chen2020simple}, with deeper layers, can capture long-distance dependency in the graph. However, it has been shown that no single model outperforms other methods across different datasets~\cite{goyal2018graph,shchur2018pitfalls}. Real-life graphs usually have a combination of several properties and they are difficult to characterize and capture by a single model. Moreover, the performance of the model has high variance with the different splits of the dataset and training configurations~\cite{shchur2018pitfalls}. In this paper, rather than using individual models, we automatically ensemble a group of effective models to exploit diverse properties of the dataset, and propose hierarchical ensemble to obtain effective and robust results.

\subsection{Automatic Neural Architecture Search for Graphs}
Neural Architecture Search (NAS) automates the process of discovering high-performance architectures and has, therefore, received broad attention~\cite{gao2020graph,nunes2020neural,zhao2020simplifying,nunes2020neural,Xu_Magicnas,Xu_NASBERT}. To automatically search a GNN model for a task, previous work mainly focuses on neural architecture search to deliver an individual model based on reinforcement learning~\cite{gao2020graph,nunes2020neural,zhao2020simplifying}, evolutionary algorithms~\cite{nunes2020neural} and one-shot supernet~\cite{zhou2019auto,anonymous2021efficient}. However, reinforcement learning and evolution based methods are computationally expensive since they require training architectures from scratch to obtain their ground-truth accuracy. One-shot supernet based methods mainly utilize weight sharing to improve training efficiency by maintaining a single copy of weights on a supernet that contains all the sub-models (architectures) in the search space and reuse the weights of shared operators from previously trained sub-models.  However, the training process with weight sharing is quite sensitive to the design of the search space~\cite{pourchot2020share,shu2019understanding,xie2020weight,yu2020train} and configurations of searching~\cite{pourchot2020share,yu2020train} due to the interference of different sub-models~\cite{zhang2020deeper}. These methods still require human expertise to carefully design the search space and searching configurations to get good results. Different from existing methods, AutoHEnsGNN automatically selects models from the candidate pool and efficiently searches the optimal configurations of the hierarchical ensemble without human intervention, which is more effective and reliable for real-life deployment. Note that the novel architectures searched by NAS algorithms are complementary to AutoHEnsGNN, that is, one can first apply NAS to search novel architectures and then add them to the candidate pool for the ensemble.

\subsection{Ensemble Learning of Graph Neural Network}
Ensemble learning, as an efficient technique that aggregates multiple machine learning models, can achieve overall high prediction accuracy and good generalization~\cite{polikar2006ensemble}. Zhao~\etal~\cite{zhao2020graph} integrate individual graph embedding for biological gene interaction prediction. Ren~\etal~\cite{ren2020paragraph} use an ensemble of GNNs to increases model accuracy for the circuit parameter prediction. Kong~\etal~\cite{kong2019forgenet} leverage tree-based feature extractors followed  by graph neural networks to learn robust representation on omics data. These methods manually selects several models for ensemble and directly apply ensemble to get more accurate and stable performance on specific tasks. Goyal~\etal~\cite{goyal2019graph} greedily add the model to the ensemble and theoretically study the benefits of the ensemble for graph tasks. Different from existing work, AutoHEnsGNN automatically and efficiently selects top-performing models, and then 
uses novel hierarchical ensemble and search algorithms for effective and robust predictions.

\section{Method}
In this section, we introduce the details of AutohEnsGNN, which automatically builds effective and robust GNN models without human intervention. We first give an overview of AutohEnsGNN, and then introduce each component of it.

\subsection{Overview}
In this paper, we focus on the semi-supervised node classification tasks in graphs, which is a fundamental problem in graph mining. We are given a graph $\mathcal{G}=(\mathcal{V},\mathcal{E},\mathbf{A})$, where $\mathcal{V}$ is the set of vertices $\{v_1,\cdots,v_n\}$, $\mathcal{E}$ is the set of edges, and $\mathbf{A}\in \mathbb{R}^{n\times n}$ is the weighted adjacency matrix of the graph. AutoHEnsGNN mainly contains two components:

\begin{itemize}[leftmargin=*]
\item Proxy evaluation for GNN model selection. Considering the heavy computation cost of training for model selection, we approximate the performance of each model on the proxy task to enable fast and parallel training. Then we can efficiently rank the candidates and select top-performing ones.
\item Hierarchical ensemble and architecture search. We first aggregate representations learned by models (same type but different initializations and layers) and then apply a weighted ensemble on different types of models. The design is based on three considerations: 1) The results of GNN models are sensitive to the initialization (see~\cite{shchur2018pitfalls}); 2) Different layers of GNN models can be aggregated to effectively exploit the information from local and global neighborhoods~\cite{xu2018representation,liu2020towards}; 3) Ensemble of models can capture different properties of graphs to further improve the performance. Then, we propose two algorithms to efficiently search configurations of the hierarchical ensemble. After searching, we can re-train the hierarchical model with optimal configurations and make predictions. The overview of AutoHEnsGNN is shown in Figure~\ref{fig:overview}.
\end{itemize}

\subsection{Proxy evaluation for GNN model Selection}\label{sec:proxy_evaluation}
Although it has been shown that various specific architectures can efficiently capture certain properties of a graph and achieve state-of-the-art results~\cite{velivckovic2018graph,kipf2016semi,hamilton2017inductive,xu2018powerful,ying2018graph,derr2018signed}, there is no model that can surpass other ones across all the datasets~\cite{shchur2018pitfalls,hu2020open}. Given a new dataset, we aim to first select models with the top performance from existing work, and then construct the ensemble for predictions. To get the accurate evaluation for a GNN model, we 1) randomly split the dataset into training and validation dataset, search the optimal configurations and train the GNN model on the training set; 2) repeat 1) many times and construct bagging of models trained on the different splits of the dataset to reduce variance in the resulting predictions. Since we cannot access the labels of the test set, we randomly split a test set from the training set. Then, we can evaluate models on the same test set and select the top ones.

Many GNN models have been proposed~\cite{zhang2020deep} and achieved state-of-the-art results on diverse domains~\cite{derr2018relevance,liu2012probabilistic,aggarwal2010managing,ying2018graph,derr2018signed,zitnik2018modeling}. It is time-consuming to accurately evaluate the performance of these candidate models and their variants. To improve the efficiency of model selection, we propose to approximate the performance of the model on the proxy task and select promising models. The proxy task contains 1) a proxy dataset: the model is trained on the sampled training dataset (sub-graph), which can largely reduce the training time and GPU memory consumption; 2) a proxy model: a model with a smaller hidden size is trained to reduce the computational cost and accelerate the training process; 3) a proxy bagging: a few models are trained on the different splits of the dataset for bagging. The key insight of proxy evaluation is that we actually care about the ranking of candidate models to select the promising models for the next stage, thus, it is not necessary to acquire highly accurate results. Due to fast training and low GPU memory consumption, we can quickly evaluate the performance of candidate models in parallel, and choose the top ones for the next stage. We empirically demonstrate the effectiveness of proxy evaluation in Section~\ref{sec:exp:rank_corr}.

\subsection{Hierarchical Ensemble and Architecture Search} \label{sec:method:hierarcihcal_ensemble}
Denote the pool of GNN models selected by proxy evaluation as $\mathcal{P}_{\text{GNN}}=\{\mathcal{N}_i\}_{i=1,\cdots,N}$, where $N$ is the number of models for ensemble. In this section, we present how to build a hierarchical ensemble and search the configurations efficiently. First, we describe the framework of the hierarchical ensemble. Then, to search the layers of each sub-model and ensemble weight, we propose AutoHEnsGNN$_{\text{Gradient}}$, which jointly optimizes architecture parameters via gradient-based search. Besides, we also propose a memory-efficient algorithm, AutoHEnsGNN$_{\text{Adaptive}}$, which optimizes each model separately and adaptively adjusts the ensemble weights based on the accuracy of models.

\subsubsection{Hierarchical Ensemble}
\begin{figure*}[!t]
\vspace{-3mm}
\centering
\includegraphics[width=0.9\textwidth]{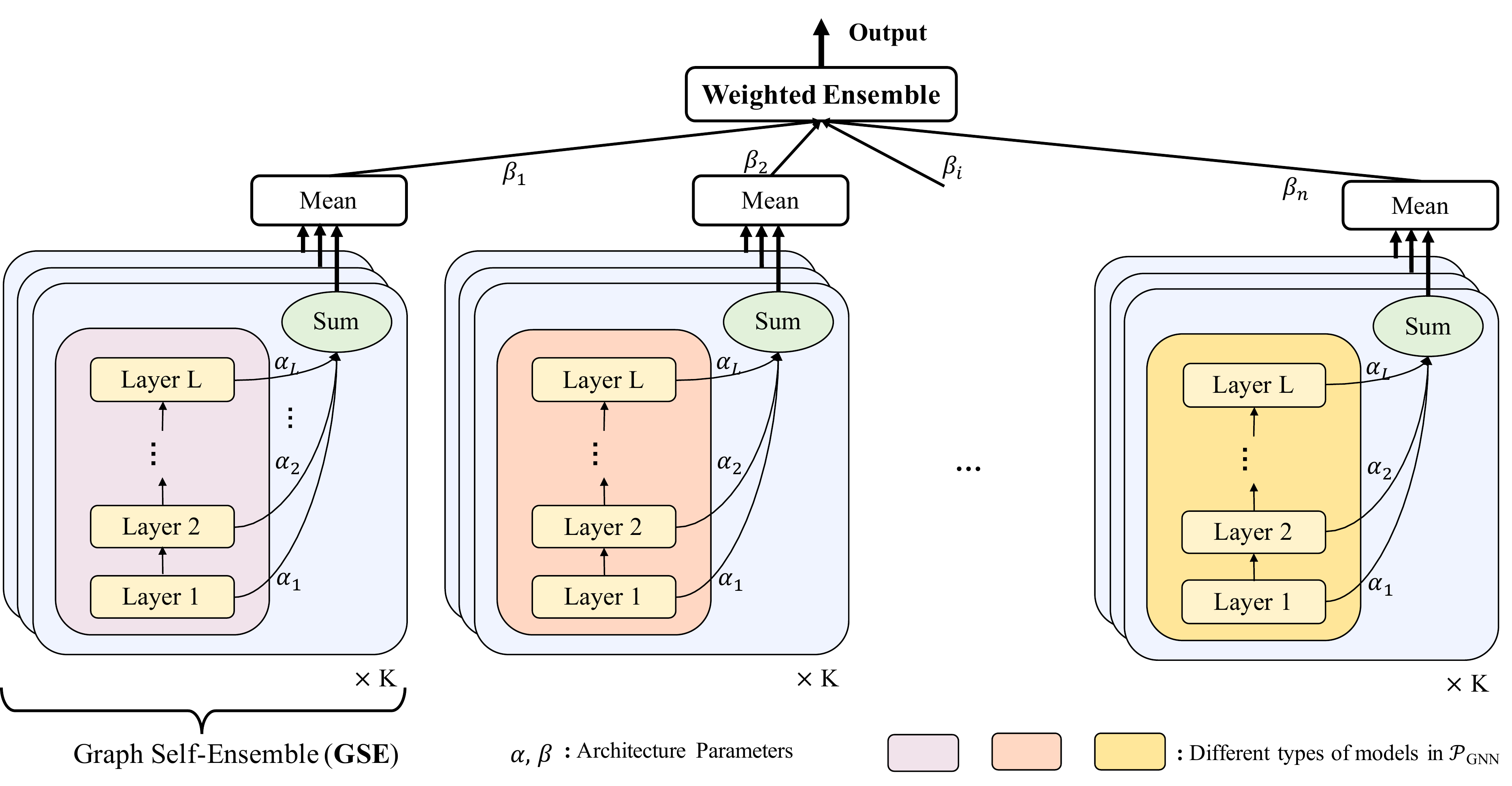}
\caption{Given $N$ different types of graph sub-models, we build a hierarchical ensemble network. For each type of sub-model, we build $K$ repeated models with different initialization seeds and layers, where each one is equipped with a one-hot vector $\alpha$ to decide which layer's output for aggregation. $K$ repeated models are aggregated (graph self-ensemble) for joint predictions to reduce the variance and efficiently exploit the information of local and global neighborhoods. Then, we further apply a weighed ensemble (weight $\beta$) of different types of sub-models to effectively learn more discriminative node presentation. $\alpha$ and $\beta$ can be searched efficiently by AutoHEnsGNN$_{\text{Adaptive}}$ and AutoHEnsGNN$_{\text{Gradient}}$ algorithms (best viewed in color).}
\label{multi_ensem}
\vspace{-1mm}
\end{figure*}

To reduce the variance and efficiently aggregate the information of local and global neighborhoods, we propose \textbf{graph self-ensemble (GSE)}, where several sub-models with different initializations and layers are assembled for joint prediction. First, for each sub-model $\mathcal{N}$ in $\mathcal{P}$,
\begin{equation}
    \mathbf{H}^{(l)}=f_{\mathcal{N}}^l(\mathbf{A},\mathbf{H}^{(l-1)}),l=1,\cdots,L
\end{equation}
where $f_{\mathcal{N}}^l$ is the $l$-th layer of network $\mathcal{N}$, $\mathbf{H}^{(0)}$ represents the input features matrix, $\mathbf{H}^{(l)}$ refers to the output of the $l$-th layer and $L$ is the maximum number of layers of the model. For node classification, the last layer of GNN predicts the labels via a \textup{softmax} classifier. Denote the predictions for $n$ nodes as $\mathbf{\hat{Y}}\in \mathbb{R}^{n\times C}$ where $C$ is number of classes. To search the number of layers for the sub-model, we further introduce a set of one-hot architecture parameters $\bm{\alpha}=\{\alpha_1,\cdots,\alpha_L\}$. Then, the prediction of the model can be written as 
\begin{equation}\label{eqn:layer_ensemble}
    \mathbf{\hat{Y}}=\text{softmax}((\sum_{l=1}^N \alpha_{l} \mathbf{H}^{(l)})\mathbf{W}),
\end{equation}
where $\mathbf{W}$ is a learnable transformation matrix. In this way, we can simply adjust $\mathbf{\alpha}$ to select the number of layers for a sub-model. Recent GNN models follow a neighborhood aggregation procedure and learn to iteratively aggregate the hidden features of every node in the graph with its adjacent nodes layer by layer~\cite{kipf2016semi,xu2018powerful,hamilton2017inductive,dwivedi2020benchmarking,gilmer2017neural,defferrard2016convolutional,velivckovic2018graph}. Theoretically, an aggregation process of $l$ layers can access the $l$-hop neighborhood. However, for the GNN model, 
\cite{liu2020towards,xu2018powerful} find that aggregating the information from different layers (local and global neighborhoods) can lead to more discriminative node representation. In light of the fact, we build several sub-models denoted as $\{\mathcal{N}^{(1)},\cdots,\mathcal{N}^{(K)}\}$ with different layers (different $\bm{\alpha}$) and the final output of the sub-model $\mathcal{N}$ can be written as
\begin{equation}\label{eqn_selfensemble}
    \mathbf{\hat{Y}_{\mathcal{N}}}=\frac{1}{K} \sum_{k=1}^{K} \mathbf{\hat{Y}_k}.
\end{equation}
The overview of GSE is shown in Fig.~\ref{multi_ensem}. $K$ models with different layers and initialization seeds are aggregated for joint predictions. 

GSE can exploit the potential of individual GNN models and reduce the variance caused by initializations. However, real-world graphs have a combination of multiple properties that are difficult to characterize by a single approach. For example, the behavior of a user in the social network may be affected by both the community (global) and its neighbors (local), which requires models to be able to find the communities and effectively aggregate the information from adjacent nodes. Thus, we first construct GSE to each sub-model in $\mathcal{P}_{\text{GNN}}$ and assemble these models as in Fig.~\ref{multi_ensem}. In this way,  self-ensemble of different layers can force each sub-model to learn different structures due to different reception fields~(layers); 2) different models may be good at extracting different features (local/global), and thus ensemble of them could combine their advantages. Formally, we can get $\mathbf{\hat{Y}_{\mathcal{N}_j}}$, according to Eqn.~\ref{eqn_selfensemble}, for each sub-model $\mathcal{N}_j$ in $\mathcal{P}_{\text{GNN}}$. Then the final results can be formulated as 

\begin{equation}
\mathbf{\hat{Y}_{\mathcal{P}}}=\sum_{j=1}^{N} \beta_{j} \mathbf{\hat{Y}_{\mathcal{N}_j}},
\end{equation}
where the different sub-models are assembled with the weight vector $\bm{\beta}=\{\beta_{1}, \cdots, \beta_{N}\}$ and $\beta_{j}=\frac{\text{exp}(\hat{\beta}_j)}{\sum_{j^{'}=1}^{N} \text{exp}(\hat{\beta}_{j^{'}})}$ is the normalized weight to determine the contribution of different sub-models to final predictions. For node classification problems, the cross-entropy loss $\mathcal{L}$ is given by
\begin{equation}
    \mathcal{L}=-\mathbf{Y} \cdot \log(\mathbf{\hat{Y}_{\mathcal{P}}}),
\end{equation}
where $\mathbf{Y}\in \mathbb{R}^{n\times C}$ is the ground truth label.

To efficiently search the architectures parameter $\bm{\alpha}$ and ensemble weight $\bm{\beta}$, we introduce AutoHEnsGNN$_{\text{Gradient}}$ and AutoHEnsGN-N$_{\text{Adaptive}}$.

\subsubsection{AutoHEnsGNN$_{\text{Gradient}}$}
Searching the effective architectures ($\bm{\alpha}$ and $\bm{\beta}$) can be formulated as a bi-level optimization problem~\cite{anandalingam1992hierarchical,colson2007overview,liu2018darts} as follows:
\begin{equation}
\begin{aligned}
\min_{\bm{\alpha},\bm{\beta}} \mathcal{L}_{val}(\bm{w}^*(\bm{\alpha},\bm{\beta}),\bm{\alpha},\bm{\beta}),\\
s.t. \qquad \bm{w}^*(\bm{\alpha},\bm{\beta})=\text{argmin}_{\bm{w}} \mathcal{L}_{train}(\bm{w}, \bm{\alpha},\bm{\beta} ).
\end{aligned}
\end{equation}
It means that, given architecture parameters $\bm{\alpha},\bm{\beta}$, the trainable weight $\bm{w}^*$ associated with the architecture are obtained by minimizing the training loss. Then our goal of architecture search is to find the best architecture configurations $\bm{\alpha}^*,\bm{\beta}^*$ that minimizes the validation loss $\mathcal{L}_{val}(\bm{w}^*,\bm{\alpha}^*,\bm{\beta}^*)$. To make the search space continous so that we can calculate the gradient, we relax the one-hot vector $\bm{\alpha}$ to a \textup{softmax} over all possible layers, and then the prediction of the sub-model in Eqn.~\ref{eqn:layer_ensemble} can be written as:
\begin{equation}
 \mathbf{\hat{Y}}=\text{softmax}((\sum_{l=1}^N \frac{\text{exp}(\hat{\alpha}_l)}{\sum_{l^{'}=1}^{N} \text{exp}(\hat{\alpha}_{l^{'}})} \mathbf{H}^{(l)})\mathbf{W}),
\end{equation}
where $\hat{\alpha}_l$ is trainable parameter. 

By relaxing $\bm{\alpha}$, we can directly calculate the gradient $\nabla_{\bm{\alpha}} \mathcal{L}_{val}(\bm{w}^*(\bm{\alpha}),\bm{\alpha})$ and $\nabla_{\bm{\beta}} \mathcal{L}_{val}(\bm{w}^*(\bm{\beta}),\bm{\beta})$. To avoid expensive inner optimization, we use the first-order approximation~\cite{liu2018darts} to iteratively update weights of model and architectures. The procedure is given by Algorithm 1. After the optimization, we can derive the sub-model with $L^*$ layers where $L^*=\text{arg}\max_{l} \frac{\text{exp}(\hat{\alpha}_l)}{\sum_{l^{'}=1}^{N} \text{exp}(\hat{\alpha}_{l^{'}})}$. Then, the layers of sub-models and $\bm{\beta}$ are fixed and each model in AutoHEnsGNN can be re-trained separately and aggregated in the way of the hierarchical ensemble.

Joint optimization of architecture parameters can search for better configurations due to consideration of the interaction of different models. Because of the gradient optimization, the search time is roughly equal to the training time of each individual model. However, the GPU memory consumption is roughly $K\times N$ times that of each individual model. To reduce the computational cost, we train the model on the proxy task (small hidden size, sampled dataset) as mentioned in Section~\ref{sec:proxy_evaluation}. 

\begin{algorithm}[h!]\label{alg:gradient_search}
\SetAlgoLined
\KwResult{Return $\bm{\alpha}$ and $\bm{\beta}$}
 i=0; Update architecture weights $\bm{\alpha}$ and $\bm{\beta}$ every $M$ iterations\;
 \While{not converged}{
  i+=1\; Get the $i$-th iteration of data\;
  Calculate the gradient $\nabla_{w} \mathcal{L}_{train}(\bm{w},\bm{\alpha},\bm{\beta})$ of weight $w$ and apply gradient descent\;
  \If{(i \% M == 0)}{
  Calculate the gradient $\nabla_{\mathbf{\bm{\alpha}}} \mathcal{L}_{val}(\bm{w}^*(\bm{\alpha}),\bm{\alpha})$ of weight $\bm{\alpha}$ and apply gradient descent\;
  Calculate the gradient $\nabla_{\bm{\beta}} \mathcal{L}_{val}(\bm{w}^*(\bm{\beta}),\bm{\beta})$ of weight $\bm{\beta}$ and apply gradient descent;
  }
 }
 \caption{Architecture Search for AutoHEnsGNN$_{Gradient}$}
\end{algorithm}

\subsubsection{AutoHEnsGNN$_{\text{Adaptive}}$}
Although the proxy task can reduce the computational cost, AutoHEnsGNN$_{\text{Gradient}}$ still consumes more GPU memory as analyzed above, which cannot meet the requirements of scenarios with very limited GPU memory. Thus, we further propose AutoHEnsGNN$_{\text{Adaptive}}$ to adaptively adjust ensemble weights according to the accuracy of each model:  1) Rather than co-training to search the number of layers of different self-ensembles, AutoHEnsGNN$_{\text{Adaptive}}$ individually optimize each one, which can largely reduce the search space from $L^{K\times N}$ to $L^K$; 2) For each self-ensemble, considering the reduced search space, AutoHEnsGNN$_{\text{Adaptive}}$ directly applies grid search to find the best configurations; 3) Rather than searching the ensemble weight, we design an adaptive $\bm{\beta}$ as follows:
\begin{equation}\label{eqn:adative_weight}
\begin{aligned}
\bm{\beta}=\text{softmax}(\frac{\textbf{acc}}{\tau}),\\
\tau=1+\frac{(1+\min(\epsilon, 1+\log(\frac{\text{\#edges}}{\text{\#nodes}}+1)))^{\lambda}}{\gamma},
\end{aligned}
\end{equation}
where $\tau$ is the annealing temperature, $\textbf{acc}=\{\text{acc}_1,\cdots,\text{acc}_N\}$ is the normalized validation accuracy of different sub-models, $\epsilon$, $\gamma$ and $\lambda$ are hyper-parameters. The intuition is that models with a high score should be assigned with a large weight. We empirically find that, when the average degree is small (sparse graph), a sharp softmax distribution can obtain better results. Given a sparse graph (the term in Eqn~\ref{eqn:adative_weight} is smaller than $\epsilon$), the sparser graph, the smaller $\tau$, which can assign more weights to promising models.

\section{Experiments and Results}
\subsection{Datasets}

\begin{table}[h!]
\caption{Statistics of anonymous public datasets  in AutoGraph Challenge of KDD Cup 2020.}
\label{table:candidate_op}
\vspace{-3mm}
\scalebox{0.95}{
\begin{tabular}{l|lllcll}
\toprule
Dataset  & \begin{tabular}[c]{@{}l@{}}Node\\ Feat.\end{tabular} &\begin{tabular}[c]{@{}l@{}}Edge \\ Feat.\end{tabular} & Directed & \multicolumn{1}{c}{\begin{tabular}[c]{@{}c@{}}Nodes\\ Train/Test\end{tabular}}  & Edges  & Classes \\
\midrule
A        &$\checkmark$ &- & - & 1088/1620  & 5278    & 7       \\
B        &$\checkmark$ & -& - & 1334/1993  & 4552    & 6       \\
C        &$\checkmark$ & -& - & 4026/5974 & 733316  & 41      \\
D        &$\checkmark$ &  $\checkmark$& $\checkmark$ & 4009/5991 & 5833962 & 20      \\
E        & - & - & - & 3011/4510  & 7804    & 3       \\
\bottomrule
\end{tabular}}
\end{table}

\subsubsection{Anonymous datasets in AutoGraph Challenge of KDD Cup 2020} The competition focuses on the semi-supervised classification tasks on graphs. As is shown in Table~\ref{table:candidate_op}, there are 5 anonymous public datasets provided by KDD Cup 2020. The datasets are collected from real-world business, and are shuffled and split into training and test parts. Each dataset contains two node files (training and test), an edge file, a feature file, a training dataset label file, and a metadata file. Table~\ref{table:data_sample} in Appendix presents an example of the anonymous dataset.

\subsubsection{Public Datasets} We also conduct experiments on three benchmark graphs Cora, Citeseer, and Pubmed~\cite{sen2008collective}, which are the most widely used benchmark datasets for semi-supervised node classification~\cite{yang2016revisiting}.
Since these graph datasets only have thousands of nodes, we further conduct experiments on ogbn-arxiv~\cite{hu2020open}, which contains 169,243 nodes, to demonstrate the scalability of our proposed methods.  We do not conduct data processing or feature engineering on these datasets.

\subsection{Analysis of Proxy Evaluation}\label{sec:exp:rank_corr}
To construct the hierarchical ensemble, firstly, we need to construct a pool of promising models by evaluating the performance of the given task. In order to avoid time-consuming evaluation, we propose to train the models on the proxy task. The key to proxy evaluation is to efficiently train the model and rank the models accurately. In this section, we use more than 20 models to demonstrate that proxy evaluation can be more memory-efficient and 5$\sim$10 $\times$ faster than accurate evaluation with high ranking accuracy.

\begin{figure*}[!t]
\centering
\begin{tabular}{cc}
\vspace{-2mm}
\includegraphics[width=0.95\textwidth]{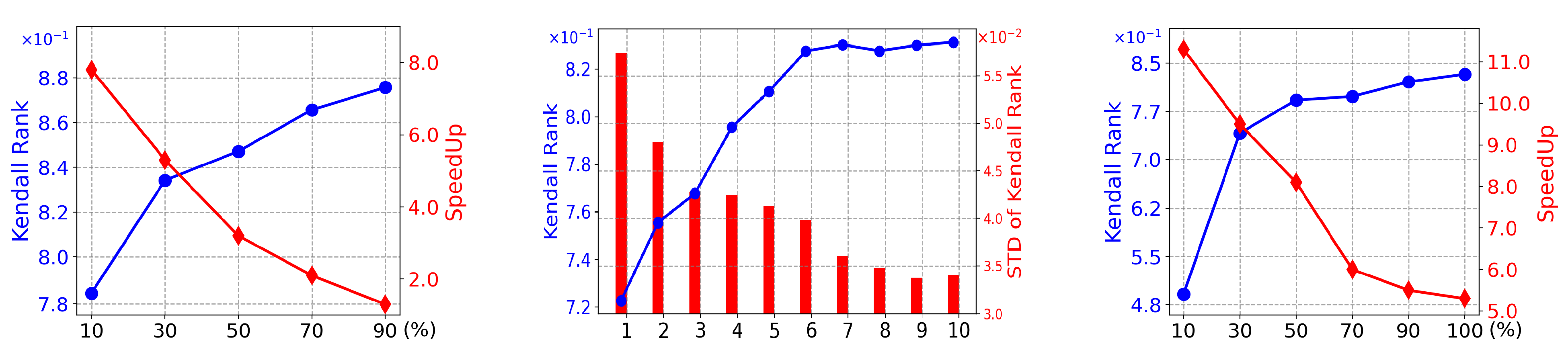}
\vspace{-2mm}\\
\begin{tabular}{lll}
  \hspace{0cm} \scriptsize $D_{\text{proxy}}$ (\%)  &
  \hspace{4.5cm}  \scriptsize $B_{\text{proxy}}$ & 
  \hspace{4.5cm}  \scriptsize $M_{\text{proxy}}$ (\%) 
\end{tabular} \vspace{0mm}
\\
(a) Dataset  A \\
\vspace{-1mm}
\includegraphics[width=0.95\textwidth]{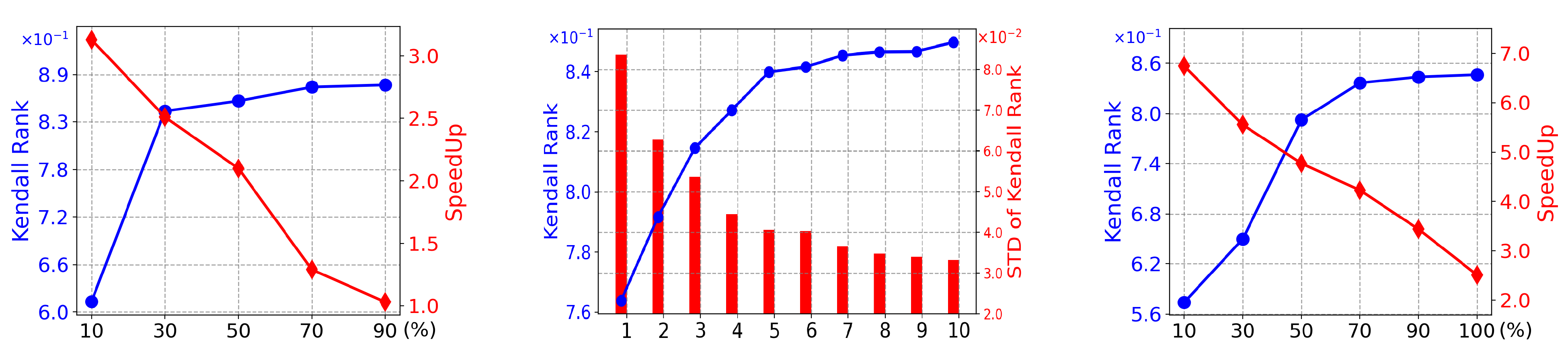}\vspace{-2mm}\\
\begin{tabular}{lll}
  \hspace{0cm}  \scriptsize $D_{\text{proxy}}$ (\%)  &
  \hspace{4.5cm}  \scriptsize $B_{\text{proxy}}$ & 
  \hspace{4.5cm}  \scriptsize $M_{\text{proxy}}$ (\%) 
\end{tabular}\vspace{0mm}\\
(b) Cora
\end{tabular}
\vspace{-1mm}
\caption{Analysis of proxy evaluation. Each row (dataset) has three sub-figures from left to right, which show the Kendall rank correlation coefficient and training efficiency related to the proxy dataset, proxy bagging, and proxy model respectively.}
\label{fig:corr_analy_a}
\end{figure*}

\subsubsection{Models for Proxy Evaluation} \label{sec:exp:proxy_eval}
To demonstrate the effectiveness of proxy evaluation, we select more than 20 types of GNN models with diverse designs of aggregators including convolutional aggregator (including spectral-based~\cite{kipf2016semi,du2017topology,defferrard2016convolutional,bianchi2019graph} and spatial-based~\cite{hamilton2017inductive,morris2019weisfeiler,xu2018powerful,wu2019simplifying,duvenaud2015convolutional,monti2017geometric,chiang2019cluster,klicpera2018predict,feng2020graph,xie2018crystal} aggregators), attention aggregator~\cite{thekumparampil2018attention,ranjan2020asap}, skip connection~\cite{fey2019just,li2020deepergcn,chen2020simple}, gate updater~\cite{li2015gated} and dynamic updater~\cite{wang2019dynamic}. These models have achieved impressive results from diverse domains and have potential to handle different types of graphs.\footnote{We just sample some models from each category in our experiments. Models that are not listed here do not indicate poor performance.} For example, \cite{morris2019weisfeiler} can make use of edge weight (edge feature) for message passing and GCNII~\cite{chen2020simple}, with deeper layers, can capture long-distance dependency in the graph.

\subsubsection{Effectiveness of Proxy Evaluation}
\paragraph{Training Configurations} \label{sec:proxy_config}
The model architectures follow the original papers and we implement them with Pytorch~\cite{NEURIPS2019_9015}. For public datasets Cora, Citeseer, and Pubmed, we use the exact training configurations and hyper-parameters as the original implements. For the anonymous dataset in AutoGraoh Challenge,  we use the same training procedure for all of the models. More details are introduced in Appendix~\ref{repro:config:proxy_eval}.

 \begin{table}[t!]
\caption{Results on the anonymous dataset in AutoGraph Challenge. P-values are calculated between AutoHEnsGNN$_{\text{Gradient}}$ and Goyal~\etal~\cite{goyal2019graph}. AutoHEnsGNN$_{\text{Gradient}}$ significantly outperforms baseline methods in A, B, C and D dataset (two-side wilcoxon test, p-values$<$0.05).}
\label{table:results_of_kddcup}
\centering
\scalebox{0.8}{
\begin{tabular}{c|ccccc}
\toprule
Method & A & B & C & D & E\\
\midrule
 GCN~\cite{kipf2016semi}    &   85.2$\pm$0.8   &  72.0$\pm$0.5  &  92.5$\pm$0.3  &  94.9$\pm$0.1  &  87.5$\pm$0.1 \\
 GAT~\cite{velivckovic2018graph}     &   83.3$\pm$0.8   &     71.2$\pm$0.4 &  89.4$\pm$0.7   & 94.6$\pm$0.4 & 87.8$\pm$0.1\\
 APPNP~\cite{klicpera2018predict}     &   76.8$\pm$0.6   &       69.2$\pm$1.0    &  72.8$\pm$0.8    & 93.8$\pm$0.0 &  87.4$\pm$0.1 \\
 TAGC~\cite{du2017topology} &  85.6$\pm$0.6  &        72.5$\pm$0.5        &   91.6$\pm$1.2    & 93.9$\pm$0.1 & 88.6$\pm$0.1 \\
 DNA~\cite{fey2019just} &  82.7$\pm$1.2  &         71.0$\pm$0.7       &   88.6$\pm$0.6    & 93.8$\pm$0.0  & 87.4$\pm$0.2 \\
 GraphSAGE~\cite{hamilton2017inductive}     &  84.4$\pm$0.7  &    71.1$\pm$0.4      &   73.6$\pm$3.5    & 93.9$\pm$0.0 & 88.1$\pm$0.2 \\
 GraphMix~\cite{verma2019graphmix}     &  86.3$\pm$0.4   &   73.1$\pm$0.7   &   87.5$\pm$0.3     & 94.0$\pm$0.2  & 87.7$\pm$0.0 \\
 Grand~\cite{feng2020graph}     &   87.0$\pm$0.5   &   73.9$\pm$0.5    &   91.0$\pm$0.7    &   93.6$\pm$0.2& 88.0$\pm$0.2\\
 GCNII~\cite{chen2020simple}     &  87.7$\pm$0.5   &  73.6$\pm$0.6  &    89.3$\pm$0.8    &   93.5$\pm$0.1 & 88.1$\pm$0.0 \\
 \midrule
 D-ensemble &   88.1$\pm$0.3   &    74.3$\pm$0.4   &   93.4$\pm$0.4  &  95.0$\pm$0.1  & 88.7$\pm$0.1\\
 L-ensemble &   88.5$\pm$0.3   &   75.0$\pm$0.2    &  93.8$\pm$0.6   &  95.5$\pm$0.1  &  88.7$\pm$0.1\\
 Goyal~\etal~\cite{goyal2019graph} &  88.7$\pm$0.3  &   74.5$\pm$0.3  &  93.9$\pm$0.4  &   95.7$\pm$0.0    & 88.7$\pm$0.1\\
 \midrule
 AutoHEnsGNN$_{\text{Ada.}}$ &  89.3$\pm$0.1   &   75.5$\pm$0.2    &  94.4$\pm$0.2  &  96.1$\pm$0.0 &  88.7$\pm$0.1 \\
 AutoHEnsGNN$_{\text{Grad.}}$ &  \textbf{89.6$\pm$0.1} &    \textbf{76.1$\pm$0.2}  &   \textbf{94.7$\pm$0.2} &  \textbf{96.3$\pm$0.0}  & \textbf{88.8$\pm$0.1}\\\bottomrule
\end{tabular}}
\vspace{-2mm}
\end{table}

\paragraph{Analysis}
We can evaluate a model accurately by 1) randomly splitting the dataset into training and validation dataset, automatically searching the best hyper-parameters and optimizing the weight of the GNN model on the train set; 2) repeating 1) many times
(10 times in our experiments) and constructing bagging of models trained on the different splits of the dataset to get robust and accurate results. To avoid time-consuming evaluation, we optimize the model with a smaller hidden size (proxy model) on the partial training set (proxy dataset) and repeat a few times of 1) for bagging (proxy bagging). We denote the ratio of the sampled training set, training times for bagging, and the ratio of the hidden size of the reduced model compared to that of the original model by $D_{\text{proxy}}$, $B_{\text{proxy}}$ and $M_{\text{proxy}}$ respectively. We calculate the Kendall rank correlation coefficient~\cite{kendall1938new} to measure the ranking correlation between models trained with proxy evaluation and accurate evaluation. The results on dataset A and Cora are presented in Figure~\ref{fig:corr_analy_a}. The left sub-figures show the Kendall rank and speedup of total training time. We can observe that proxy dataset with $D_{\text{proxy}}$=30\% can achieve a good trade-off of ranking accuracy (0.836 on Dataset A and 0.841 on Cora) and speed (4.7$\times$ on Dataset A and 2.6$\times$ on Cora). In the middle sub-figure, we fix $D_{\text{proxy}}$=30\% and study the effect of proxy bagging. Since bagging is used mainly to reduce variance, we also report the STD of Kendell Rank by repeating experiments 100 times. From Figure~\ref{fig:corr_analy_a}, we can find that $B_{\text{proxy}}$=6 can get both relatively high Kendall rank and low variance. Furthermore, we fix $D_{\text{proxy}}$=30\% and $B_{\text{proxy}}$=6 to study the $M_{\text{proxy}}$ in the right sub-figures. It can be seen that Kendall rank is low when the model is small ($M_{\text{proxy}}$=10\%), a reasonable Kendall rank (0.758 on Dataset A and 0.795 on Cora) and speedup (10.4$\times$ on Dataset A and 5.7$\times$ on Cora) can be achieved by  $M_{\text{proxy}}$=50\%. $M_{\text{proxy}}$=50\% can also reduce GPU memory consumption. These results demonstrate that, by combining proxy dataset, bagging, and model, we can greatly improve the efficiency of evaluation and keep a high Kendall rank. Due to the GPU memory reduction, we can also conduct a parallel evaluation of different models (up to the limit of GPU memory) to further accelerate this procedure. In the following parts, we set $D_{\text{proxy}}$=30\%, $B_{\text{proxy}}$=6, and $M_{\text{proxy}}$=50\% as default for proxy evaluation.

\subsection{Results of AutoHEnsGNN}\label{exp:node_class}
\paragraph{Training Configurations} For each dataset, AutoHEnsGNN can automatically run the proxy evaluation and select top models $\mathcal{P}_{\text{GNN}}$ for the hierarchical ensemble. As described in Section~\ref{sec:method:hierarcihcal_ensemble}, AutoHEnsGN-N$_{\text{Gradient}}$ and AutoHEnsGNN$_{\text{Adaptive}}$ are proposed to search $\alpha$ and $\beta$. Detailed hyper-parameters of search are presented in Appendix~\ref{repro:config:autohensgnn}. After searching, each sub-model is re-trained separately and aggregated in the way of the hierarchical ensemble. For the anonymous datasets in AutoGraph Challenge, we construct bagging by randomly splitting the dataset twice for all methods. For the public datasets, we do not construct bagging on the different splits of the dataset and follow standard fixed training/validation/test split (20 nodes per class for training, 500 nodes for validation and 1,000 nodes for test)~\cite{yang2016revisiting} in Cora, Citeseer and Pubmed for a fair comparison with other methods.

\paragraph{Results on Anonymous datasets in AutoGraph Challenge}
In Table~\ref{table:results_of_kddcup}, we present top models in AutoHEnsGNN $\mathcal{P}_{\text{GNN}}$ and add the ensemble-based baselines for comparison. The ensemble-based methods are as follows:
\begin{itemize}
\item \textbf{D-ensemble}  directly averages the scores of different models for predictions.
\item \textbf{L-ensemble} stacks a learnable weight for ensemble and learns the ensemble weight on the validation set. 
\item Goyal~\etal~\cite{goyal2019graph} greedily adds the hidden representation learned by the next model to the current representation for the ensemble.
\end{itemize}

These ensemble methods share the same GNN pools $\mathcal{P}_{\text{GNN}}$ with AutoHEnsGNN for a fair comparison. In Table~\ref{table:results_of_kddcup}, AutoHEnsGNN$_{\text{Adaptive}}$ outperform all baselines on five datasets. Furthermore,  AutoHEnsGN-N$_{\text{Gradient}}$, which considers the relationship of different models by jointly training multiple graph self-ensembles to search the optimal configurations, can get better results than AutoHEnsGNN$_{\text{Adaptive}}$. We can observe that, the performance of models on dataset C differs significantly. Ensemble-based methods (Ensemble, L-ensemble, Goyal~\etal~\cite{goyal2019graph} and AutoHEnsGNN) using the models selected by proxy evaluation can get better results than individual models, which shows that proxy evaluation can select top models for the ensemble. Furthermore, AutoHEnsGNN equipped with the hierarchical ensemble can achieve a lower standard deviation, which demonstrates the robustness of our methods. 

\begin{table}[h!]
\vspace{-2mm}
\caption{Results on the Cora, Citeceer and Pubmed. P-values are calculated between AutoHEnsGNN$_{\text{Gradient}}$ and L-ensemble. AutoHEnsGNN$_{\text{Gradient}}$ significantly outperforms baseline methods on all datasets (p-values$<$0.05).}
\label{table:results_of_public}
\centering
\begin{tabular}{c|ccc}
\toprule
Method & Cora & \multicolumn{1}{c}{Citeseer} & Pubmed   \\
\midrule
 GCN~\cite{kipf2016semi}    &    81.5  &               70.3               &   79.0    \\
 GAT~\cite{velivckovic2018graph}     &    83.0$\pm$0.7  &         72.5$\pm$0.7         &   79.0$\pm$0.3       \\
 APPNP~\cite{klicpera2018predict}     &    83.8$\pm$0.3  &         71.6$\pm$0.5         &   79.7$\pm$0.3       \\
 Graph U-Net~\cite{gao2019graph} &  84.4$\pm$0.6  &         73.2$\pm$0.5         &   79.6$\pm$0.2       \\
 SGC~\cite{wu2019simplifying}     &    82.0$\pm$0.0  &         71.9$\pm$0.1         &   78.9$\pm$0.0      \\
 MixHop~\cite{abu2019mixhop}     &    81.9$\pm$0.4  &         71.4$\pm$0.8         &   80.8$\pm$0.6       \\
 GraphSAGE~\cite{hamilton2017inductive}     &    78.9$\pm$0.8  &         67.4$\pm$0.7         &   77.8$\pm$0.6      \\
 GraphMix~\cite{verma2019graphmix}     &    83.9$\pm$0.6  &         74.5$\pm$0.6        &   81.0$\pm$0.6      \\
 GRAND~\cite{feng2020graph}     &    85.4$\pm$0.4  &         75.5$\pm$0.4         &   82.7$\pm$0.6      \\
 GCNII~\cite{chen2020simple}     &    85.5$\pm$0.5  &         73.4$\pm$0.6         &   80.2$\pm$0.4      \\
 \midrule
 D-ensemble &   85.6$\pm$0.3   &      75.7$\pm$0.2           &   82.7$\pm$0.4   \\
 L-ensemble &  85.9$\pm$0.2   &      76.0$\pm$0.2           &  82.9$\pm$0.1     \\
 Goyal~\etal~\cite{goyal2019graph} &    85.9$\pm$0.3   &     75.7$\pm$0.2            &    82.8$\pm$0.2
   \\
  MixCobra~\cite{fischer2019aggregation} &   85.3$\pm$0.7    &   75.5$\pm$0.7          &   82.3$\pm$0.4
   \\
 \midrule
 AutoHEnsGNN$_{\text{Adaptive}}$ &   86.1$\pm$0.2   &        76.3$\pm$0.1          &   83.5$\pm$0.2   \\
 AutoHEnsGNN$_{\text{Gradient}}$ &   \textbf{86.5$\pm$0.2}
   &          \textbf{76.9$\pm$0.2}       &  \textbf{84.0$\pm$0.1}    \\\bottomrule
\end{tabular}
\end{table}

\begin{figure*}[ht]
\vspace{2mm}
\centering
\begin{tabular}{cc}
\includegraphics[width=\textwidth]{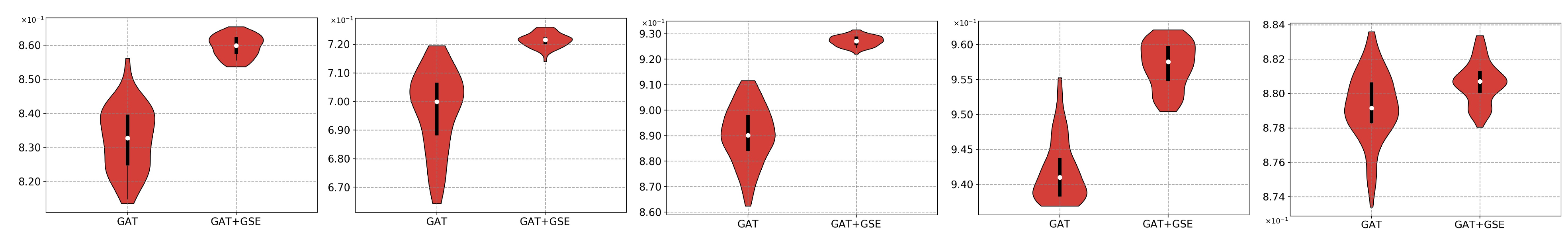}\vspace{-0mm}\\
\begin{tabular}{lllll}
  \hspace{0.4cm} \small Dataset A  &
  \hspace{1.8cm} \small Dataset B & 
  \hspace{1.8cm} \small Dataset C & 
  \hspace{1.8cm} \small Dataset D & 
  \hspace{1.8cm} \small Dataset E 
\end{tabular}
\end{tabular}
\caption{Analysis of model variance with different initializations. ``+GSE'' refers to using graph self-ensemble.}
\label{fig:robustness_study_gse}
\vspace{-2mm}
\end{figure*}

\begin{figure*}[ht]
\vspace{0mm}
\begin{tabular}{cc}
\includegraphics[width=1.0\textwidth]{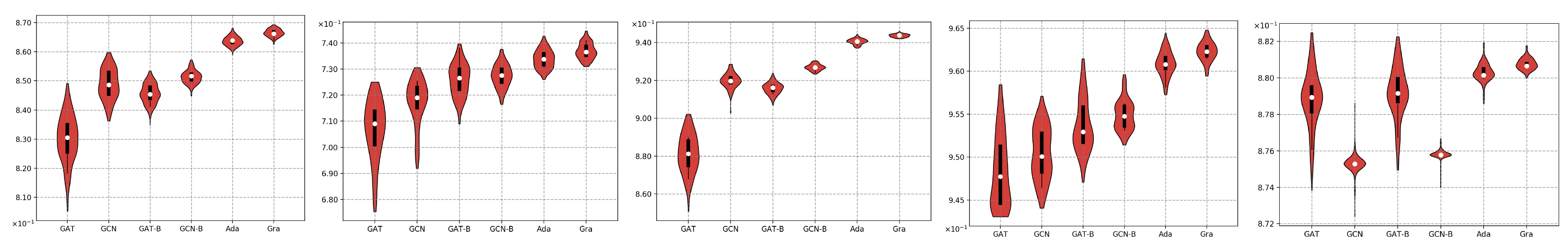}\\
\begin{tabular}{lllll}
  \hspace{0.4cm} \small Dataset A  &
  \hspace{1.8cm} \small Dataset B & 
  \hspace{1.8cm} \small Dataset C & 
  \hspace{1.8cm} \small Dataset D & 
  \hspace{1.8cm} \small Dataset E 
\end{tabular}
\end{tabular}
\caption{Analysis of model variance on different splits of datasets. ``-B'' refers training with bagging. ``Ada'' and ``Gra'' refer to AutoHEnsGNN$_{\text{Adaptive}}$ and AutoHEnsGNN$_{\text{Gradient}}$.}
\label{fig:robustness_study_hir}
\vspace{-1mm}
\end{figure*}

\paragraph{Results on the public datasets} We compare AutoHEnsGNN with 1) start-of-the-art models~\cite{kipf2016semi,velivckovic2018graph,klicpera2018predict,gao2019graph,wu2019simplifying,abu2019mixhop,hamilton2017inductive,verma2019graphmix,feng2020graph,chen2020simple} and 2) ensemble-based methods including D-ensemble, L-ensemble, Goyal~\etal~\cite{goyal2019graph} and kernel-based ensemble baseline MixCobra~\cite{fischer2019aggregation}. As shown in Table~\ref{table:results_of_public}, among recent start-of-the-art single models, GCNII~\cite{chen2020simple} achieves 85.5\% accuracy on Cora dataset and GRAND~\cite{feng2020graph} obtains 75.5\% and 82.7\% accuracy on Citeseer and Pubmed. Compared to GCNII and GRAND, AutoHEnsGNN consistently achieves large-margin outperformance across all datasets with low standard deviation, which demonstrates the effectiveness and robustness of our proposed methods. When compared to ensemble-based methods~\cite{goyal2019graph}, the proposed hierarchical ensemble achieves 0.6\%, 1.2\%, and 1.2\% (absolute differences) improvements on Cora, Citeseer, and Pubmed respectively.

\begin{table*}[h!]
\vspace{-1mm}
\centering
\caption{Ablation Study. ``PE'' and ``GSE'' refer to proxy evaluation and graph self-ensemble.}
\label{table:results_of_ablation}
\begin{tabular}{c|ccccc}
\toprule
Method & A & B & C & D & E \\
\midrule
 Single Model      &    65.2$\sim$87.7    & 42.0$\sim$74.5   &   29.1$\sim$92.5 & 93.1$\sim$95.0 & 46.5$\sim$88.2
      \\
 Random Ensemble    &   83.3$\pm$2.5   &  73.5$\pm$1.3
   &      79.4$\pm$4.7   & 94.2$\pm$0.3
 &  87.7$\pm$0.2
 \\\midrule
 Ensemble + PE     &    87.3$\pm$0.8   & 74.7+-0.3
  &      93.3$\pm$0.9 & 95.3$\pm$0.1
 &    88.0$\pm$0.1
  \\
 Ensemble + PE + GSE  &   88.6$\pm$0.3   & 75.0$\pm$0.4
  &   93.8$\pm$0.2  & 95.7$\pm$0.1 & 88.3$\pm$0.1 \\
 Ensemble + PE + GSE + Adaptive   &   89.3$\pm$0.1   &   75.5$\pm$0.2    &  94.4$\pm$0.2  &  96.1$\pm$0.0 &  88.7$\pm$0.1  \\
 Ensemble + PE + GSE + Gradient   &  \textbf{89.6$\pm$0.1} &    \textbf{76.1$\pm$0.2}  &   \textbf{94.7$\pm$0.2} &  \textbf{96.3$\pm$0.0}  & \textbf{88.8$\pm$0.1} \\\bottomrule
\end{tabular}
\vspace{-3mm}
\end{table*}

\subsection{Method Analysis}
\subsubsection{Robustness Study}\label{sec:robustness}
To demonstrate the variance caused by different initializations, we present the results of GAT~\cite{velivckovic2018graph} in Figure~\ref{fig:robustness_study_gse} by repeating 100 experiments on a fixed split of the dataset. From the figure, for example, we can find that the results of GAT vary from 81.2\% to 85.5\% on dataset A and 86.2\% to 91.1\% on dataset C with only different initializations, which is undesirable for practical deployment. By using GSE ($K=3$), the 
differences of GAT due to initializations are reduced from 4.3\% to 1.1\% on dataset A and 4.9\% to 1.0\% on dataset C. Besides, it can be seen that the average performance is improved significantly. Since GSE can adaptively aggregate models with different layers, models equipped with GSE can exploit the diverse information from different neighborhoods and learn more discriminative and robust representation.  

We further analyze the variance of training on different training/validation splits. We first run GAT~\cite{velivckovic2018graph} and GCN~\cite{kipf2016semi} with 100 random training/validation splits. As shown in Figure~\ref{fig:robustness_study_hir}, training on different data splits is also a factor leading to high variance. For example, GCN~\cite{kipf2016semi} vary from 69.2\% to 73.1\% on dataset B. Since the nodes in the graph are not mutually independent, different splits of training sets may lead to that models fit different data distributions and have different generality on the test set. When leveraging bagging on different splits of datasets (three splits in the experiments), we can see that the differences caused by data splits are reduced from 3.9\% to 2.0\% for GCN on dataset B. Similarly, GAT~\cite{velivckovic2018graph} and results on other datasets share the same case. We further adopt GAT and GCN as the candidate pools, build AutoHEnsGNN with three splits for bagging, and repeat 100 experiments. As shown in Figure~\ref{fig:robustness_study_hir}, it can be clearly observed that AutoHEnsGNN can achieve better results with a lower variance, which demonstrates the effectiveness and robustness of our models.

\subsubsection{Ablation Study} The proposed AutoHEnsGNN consists of several key components: PE (proxy evaluation), GSE (graph self-ensemble), and search algorithms (adaptive or gradient-based). To study the contribution of each component of AutoHEnsGNN, the performance of successively adding each component is presented in Table~\ref{table:results_of_ablation}. The results indicate that all of these procedures are crucial. Specifically, compared to the random ensemble (random select models for the ensemble), the ensemble with models selected by PE can largely improve the performance, demonstrating the necessity of selecting the proper models. By adding GSE, the model consistently achieve the lower variance and better accuracy across all datasets. Furthermore, models via the adaptive and gradient-based algorithms, 
which efficiently and effectively search configurations of the hierarchical ensemble, can obtain better results.

\subsubsection{Scalability Study}
To investigate the scalability of AutoHEnsGNN, we conduct experiments on the large citation dataset \emph{ogbn-arxiv}~\cite{hu2020open}. Ogbn-arxiv is a recently proposed large-scale dataset of paper citation networks~\cite{hu2020open} with 169,343 nodes and 1,166,243 edges. In the citation networks, nodes represent arXiv papers and edges denote the citations between two papers. We follow the public data split~\cite{hu2020open} to train models for a fair comparison.

Since AutoHEnsGNN$_{\text{Adaptive}}$ individually optimizes each graph self-ensemble (GSE) and directly searches the best configurations, it consumes the same GPU memory as baseline graph neural networks. However, AutoHEnsGNN$_{\text{Gradient}}$ leverages joint gradient optimization to search for the best architecture parameters and ensemble weights, which requires more GPU resources. To reduce the GPU memory consumption, in the search stage, AutoHEnsGNN$_{\text{Gradient}}$ uses the proxy model and proxy dataset (as mentioned in Section~\ref{sec:proxy_evaluation}). The configurations of the proxy model and proxy dataset are the same as those mentioned in Section~\ref{sec:proxy_config}. We empirically find that proxy dataset and evaluation for AutoHEnsGNN$_{\text{Gradient}}$ do not degrade its final re-training performance for such a large dataset. As a result, both AutoHEnsGNN and baselines can be run in one NVIDIA P40 GPU.

\begin{table}[h!]
\centering
\caption{Results on the test set of large dataset ogbn-arxiv. ``NormAdj'' refers to the normalized adjacency matrix, ``C\&S'' refers to Correct and Smooth and ``Label Reuse'' refers to utilizing not only the true values on training nodes, but also predicted values on test nodes as inputs~\cite{huang2020combining,wang2021bag}. P-values are calculated between AutoHEnsGNN$_{\text{Gradient}}$ and Goyal~\etal~\cite{goyal2019graph}. AutoHEnsGNN$_{\text{Gradient}}$ significantly outperforms baseline methods (p-values$<$0.05).}
\label{table:results_of_ogb_arxiv}
\begin{tabular}{c|ccc}
\toprule
Method & obgn-arxiv   \\
\midrule
MLP~\cite{hu2020open}    &    57.7$\pm$0.1     \\
Node2vec~\cite{grover2016node2vec}    &    71.3$\pm$0.1      \\
 GCN~\cite{kipf2016semi}    &    71.7$\pm$0.3      \\
 GAT~\cite{velivckovic2018graph}     &    73.2$\pm$0.2     \\
 APPNP~\cite{klicpera2018predict}     &   66.4$\pm$0.4     \\
 SIGN~\cite{sign_icml_grl2020} &        72.0$\pm$0.1 \\
 SGC~\cite{wu2019simplifying}     &     69.4$\pm$0.2     \\
 GaAN~\cite{zhang2018gaan}     &   72.0$\pm$0.2   \\
 GraphSAGE~\cite{hamilton2017inductive}     &   71.5$\pm$0.3     \\
 JKNet~\cite{xu2018representation}     &   72.2$\pm$0.2  \\
 DAGNN~\cite{liu2020towards}     &     72.1$\pm$0.3 \\
 DeeperGCN\cite{li2020deepergcn} & 71.9$\pm$0.2 \\
 GCNII~\cite{chen2020simple}     &    72.7$\pm$0.2    \\
 \midrule
 GAT + C\&S~\cite{huang2020combining} &      73.9$\pm$0.1 \\
 GAT + NormAdj + Label reuse + C\&S~\cite{huang2020combining} &      73.9$\pm$0.1 \\
 RevGAT + NormAdj + Label reuse~\cite{li2021training} &      74.0$\pm$0.0 \\
 \midrule
 D-ensemble &       73.9$\pm$0.0   \\
 L-ensemble &       74.0$\pm$0.2 \\
 Goyal~\etal~\cite{goyal2019graph}&   74.0$\pm$0.1    &     
   \\
 \midrule
 AutoHEnsGNN$_{\text{Adaptive}}$ &   74.2$\pm$0.0    \\
 AutoHEnsGNN$_{\text{Gradient}}$ &   \textbf{74.3$\pm$0.0} \\\bottomrule
\end{tabular}
\vspace{-2mm}
\end{table}

The results on the ogbn-arxiv dataset are shown in Table~\ref{table:results_of_ogb_arxiv}. The baseline models are implemented following their publicly available codes. Previous work~\cite{huang2020combining,wang2021bag} enhances their models with additional tricks such NormAdj, label reuse and C\&S~\cite{huang2020combining,wang2021bag} to boost the performance. For a fair comparison, we also apply them on ensemble-based baselines and our AutoHEnsGNN. From Table~\ref{table:results_of_ogb_arxiv}, we can see that AutoHEnsGNN$_{\text{Gradient}}$ outperforms baseline models and ensemble-based models with higher accuracy 74.3\% score and lower variance for the large dataset ogbn-arxiv, which demonstrates the scalability of our proposed methods.

\begin{table*}[hbt!]
\caption{Runtime statistics on large dataset ogbn-arxiv. The ``Peak GPU'' means the max GPU usage during the training process. ``PE'' refers to proxy evaluation.}
\label{table:runtime_statistics}
\centering
\begin{tabular}{c|cc|cc|cc|cc}
\toprule
\multirow{3}{*}{Model} & \multicolumn{2}{c|}{Model Selection}              & \multicolumn{4}{c|}{Training}                             & \multicolumn{2}{c}{Total}                         \\ \cline{2-9} 
                       & \multirow{2}{*}{Time (s)} & \multirow{2}{*}{Peak GPU} & \multicolumn{2}{c|}{Time (s)} & \multicolumn{2}{c|}{Peak GPU} & \multirow{2}{*}{Time (s)} & \multirow{2}{*}{Peak GPU} \\ \cline{4-7}
                       &                       &                           & Search      & Train       & Search         & Train        &                       &                           \\ \cline{1-9} 
AutoHEnsGNN$_{\text{Adaptive}}$               &    12410                   &          10.2G                 & 511    & 8989     &  2.8G          & 2.6G          &           21910            &             10.2G              \\
AutoHEnsGNN$_{\text{Gradient}}$               &    12410                   &             10.2G              & 696       & 8121      & 6.9G          & 2.5G          &          21227             &           10.2G                \\
D-Ensemble, L-Ensemble &        12410               &   10.2G                        & -        & 10116      & -           & 2.6G          &           22526            &       10.2G                    \\
Goyal~\etal~\cite{goyal2019graph}                  &    12410                   &        10.2G                   &  -        & 10116      & -           & 2.6G          &        22526               &            10.2G               \\
Ensemble + PE          &        12410               &     10.2G                      & -        & 3293      & -           & 2.6G          &            14266           &      10.2G                     \\
Ensemble               &        52730               &    19.4G                       & -        & -      & -           & -          &           52730            &    19.4G  \\\bottomrule                    
\end{tabular}
\vspace{-2mm}
\end{table*}

\subsubsection{Runtime Statistics}
The experiments on Section IV-D show that a single model suffers from high variance and undesirable performance, which is also discussed by Shchur et al.~\cite{shchur2018pitfalls} and is named pitfalls of GNN evaluation. It hinders practical usage in real-life scenarios for a single GNN model. To obtain robust and accurate predictions, AutoHEnsGNN leverages hierarchical ensemble on a pool of effective models to reduce variance and improve accuracy, and thus surpass ensemble baselines. In addition to superior performance, we further investigate runtime statistics. We conduct experiments on the large graph dataset ogbn-arxiv and make a comparison with other ensemble methods in Table~\ref{table:runtime_statistics}. ``Ensemble'' means the naive ensemble of all possible candidate models (20 models as introduced in Section~\ref{sec:exp:proxy_eval}).
``Ensemble+PE'' refers to the ensemble of the models in the pool selected by proxy evaluation. Other methods adopt the proxy evaluation and ensemble models with different initialization to reduce variance. For example, in L-ensemble, each kind of model ($N$ models in the pool) is trained with $K$ different initialization. Then L-ensemble learns the ensemble weights of the $N\times K$ models ($N$=3, $K$=3 for ogbn-arxiv). The statistics are measured on a single NVIDIA P40 GPU. 

From Table~\ref{table:runtime_statistics}, we can observe that 1) Proxy evaluation can greatly improve the training efficiency in terms of time and GPU memory; 2) ``Ensemble+PE'' obtains the lowest time cost since it does not run the models multiple times with different initialization. Despite its time efficiency, it usually cannot achieve good performance as discussed in Section IV-D. As reported in experiments on dataset A in Table V, ``Ensemble+PE'' only reaches accuracy 87.3 with a variance of 0.8. By applying our proposed graph self-ensemble and searching algorithms, the accuracy can be improved by a large margin (87.3 to 89.6) with lower variance (0.1); 3) Other methods consume similar time and GPU memory in total, in which AutoHEnsGNN$_{\text{Gradient}}$ uses slightly less time; 4) AutoHEnsGNN$_{\text{Gradient}}$ consumes more GPU memory and less time cost than AutoHEnsGNN$_{\text{Adaptive}}$ at the training stage. For scenarios where models should be trained in an online manner within limited GPU memory,  AutoHEnsGNN$_{\text{Adaptive}}$ can be a better choice.

\begin{figure}[h!]
\centering
\includegraphics[width=0.5\textwidth]{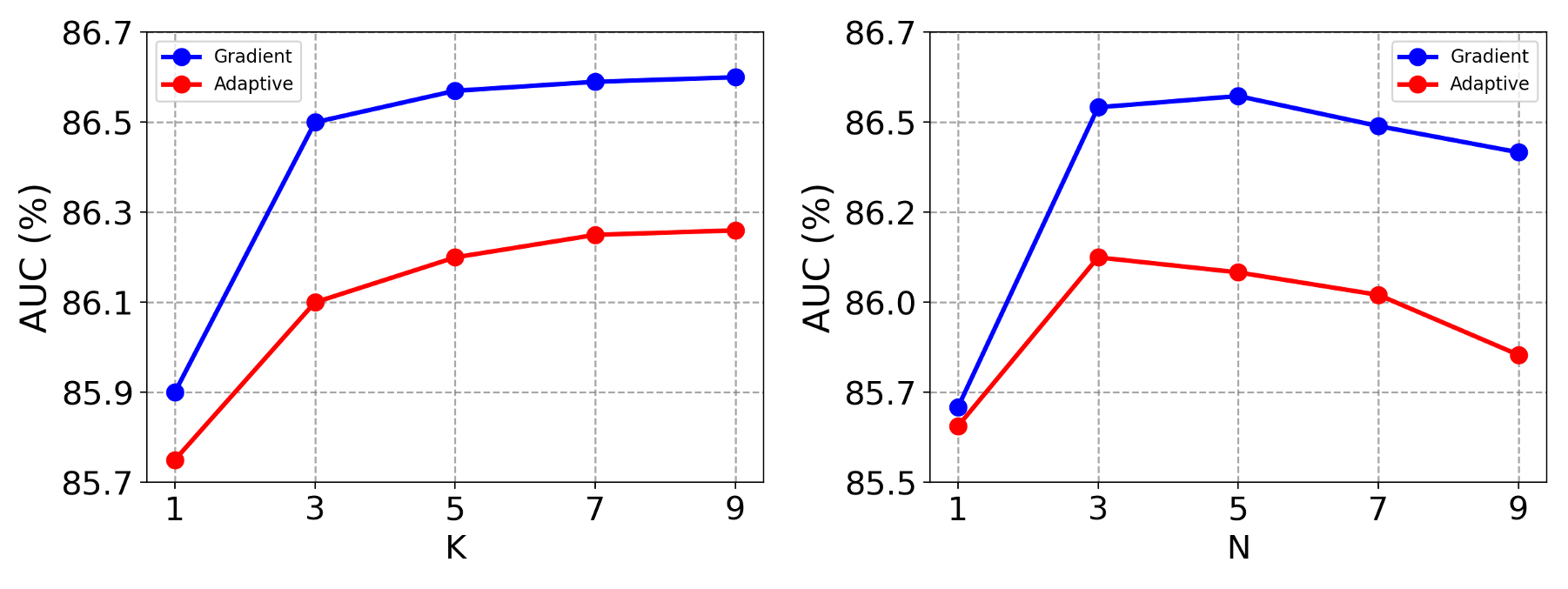}
\vspace{-7mm}
\caption{Hyper-parameters study on Cora for the hierarchical ensemble. $N$ refers to pool size, and $K$ refers to the number of models for self-ensemble. Values in the figure are the average of 10 repeated experiments with different seeds.} 
\label{fig:k_n_study}
\vspace{-3mm}
\end{figure}

\begin{figure}[h!]
\centering
\includegraphics[width=0.5\textwidth]{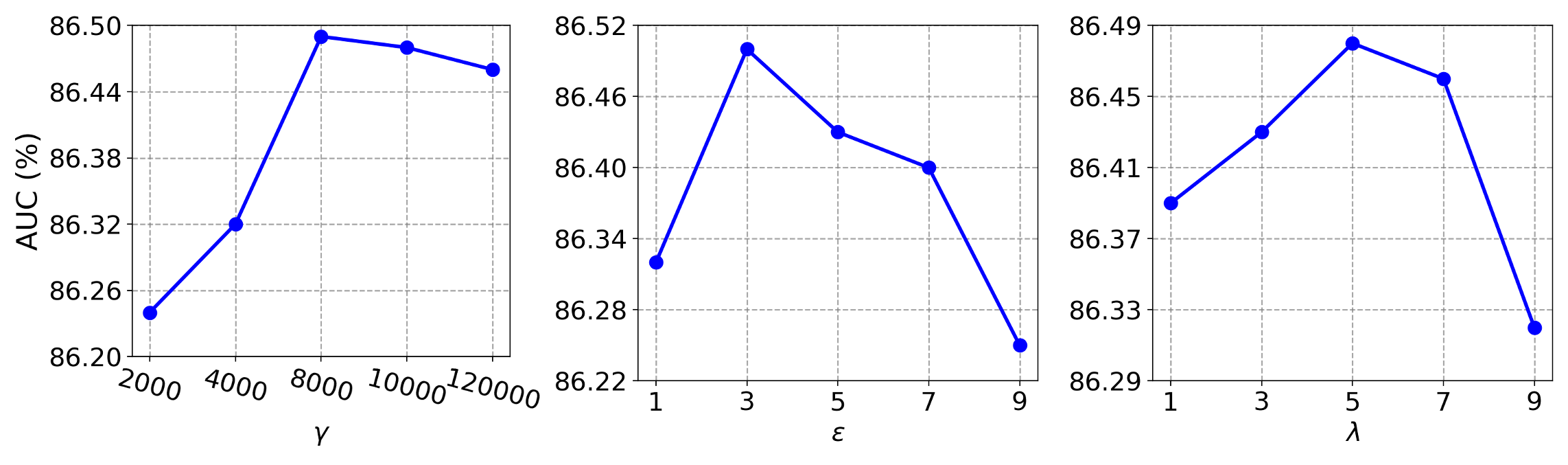}
\vspace{-7mm}
\caption{Hyper-parameters study on Cora for AutoHEnsGNN$_{\text{Adaptive}}$. $\lambda$, $\epsilon$ and $\gamma$ decide the annealing temperature in Eqn~(8). Values in the figure are the average of 10 repeated experiments with different seeds.}
\label{fig:adaptive_study}
\vspace{-3mm}
\end{figure}

\subsubsection{Hyper-parameter study}
We study the effects of hyper-parameters in AutoHEnsGNN on Cora dataset. The experiments are designed by adjusting a single hyper-parameter while keeping others fixed (using the default values).
\paragraph{$K$ and $N$} There are two hyper-parameters in the hierarchical ensemble: the pool size $N$ and the number of models $K$ for self-ensemble. We first present a study on the pool size by varying $N$ from 1 to 9. As shown in Figure~\ref{fig:k_n_study}, a large pool size cannot bring more benefits since some models with poor performance may be selected, and a small pool size such as $N=3$ is better. 
We further study the number of models $K$ for self-ensemble by varying it from 1 to 9. We can find that more models with different initialization for ensemble can lead to more robust and better performance. Considering the trade-off of efficiency and performance, $K=3$ is set as the default value in our experiments. In all, we can observe that the performance of AutoHEnsGNN is relatively stable across different hyper-parameters.

\paragraph{$\epsilon$, $\gamma$ and $\lambda$} For AutoHEnsGNN$_{\text{Adaptive}}$, three hyper-parameters decide the adaptive ensemble weight $\bm{\beta}$. The study of these hyper-parameters is shown in Table~\ref{fig:adaptive_study}. Small $\lambda$, $\epsilon$ or large $\gamma$ can lead to bias to those models with high accuracy, while large $\lambda$, $\epsilon$ or small $\gamma$ encourage to assign ensemble weights evenly. According to experimental results, proper hyper-parameters, which can incorporate model diversity and meanwhile assign higher weights on high-accuracy models, can achieve best performance.

\subsection{Automatic Prediction without Human Intervention on the
AutoGraph Challenge of KDD Cup 2020}\label{sec:kddcup_results}
The competition requires participants to automatically make predictions on the final 5 real-life datasets without any human intervention. These tasks are constrained by the time budget and GPU memory. To avoid failure caused by unexpected GPU memory/time consumption, we submit the codes implemented with AutoHEnsGNN$_{\text{Adaptive}}$ and reduce the search space of $\alpha$ and other hyper-parameters (e.g., dropout rate)\footnote{The code is publicly available at

\url{https://github.com/aister2020/KDDCUP\_2020\_AutoGraph\_1st\_Place.}}. As is shown in Table~\ref{exp:kddcup}, our team aister won 1st place, which achieves the best rank score on average of 5 datasets in the final phase. The competition results on the 5 real-life datasets demonstrate the effectiveness of AutoHEnsGNN.

\begin{table}[h!]
\centering
\caption{Results of KDD Cup 2020 AutoGraph Competitions. Our team ``aister'' won 1st place with the best rank score in the final phase.}
\label{exp:kddcup}
\begin{tabular}{llc}
\toprule
Rank                 & Team & Average Rank Score     \\\midrule
1   & \textbf{aister (ours)} &  \textbf{4.8}   \\
2   & PASA\_NJU & 5.2    \\
3   & qqerret & 5.4    \\
4   & common &  6.6   \\
5   & PostDawn & 7.4    \\
6   & SmartMN-THU & 7.8    \\
7   & JunweiSun &  7.8   \\
8   & u1234x1234 &  9.2  \\
9   & shiqitao &  9.6   \\
10   & supergx &  11.8   \\
\bottomrule
\end{tabular}
\vspace{-2mm}
\end{table}

\section{Edge Prediction and Graph Classification}

To investigate the generality of AutoHEnsGNN, we further conduct experiments on two common graph tasks: edge prediction and graph classification. The former is conducted on three citation networks: Cora, CiteSeer and PubMeb~\cite{sen2008collective}, and the latter is conducted chemical compounds dataset PROTEINS~\cite{morris2020tudataset}. PROTEINS is a binary classification dataset, which contains 1,113 graphs with average degrees 39.06 and average links 72.82. The data splits follow their common practice~\cite{sen2008collective, morris2020tudataset}. We leverage 6~\cite{pan2021neural, davidson2018hyperspherical,di2020mutual,mavromatis2020graph,yang2018binarized,pan2018adversarially} and 7~\cite{xu2018powerful,hamilton2017inductive,nouranizadeh2021maximum,nguyen2019universal,di2020mutual,gao2019graph,zhang2019hierarchical} GNN models as the candidate pool for graph prediction and classification task respectively. The K is set to 3 and N is set to 2. Other hyper-parameters are the same as those for node classification tasks in Section~\ref{exp:node_class}.

As shown in Table~\ref{table:link_prediction} and Table~\ref{table:graph_prediction}, AutoHEnsGNN consistently outperforms other methods by a large margin, which verifies the generality of our algorithms. Meanwhile, AutoHEnsGNN greatly surpasses the existing work in edge prediction tasks and achieves the state-of-the-art results\footnote{See \url{https://paperswithcode.com/task/link-prediction}.}.

\begin{table}[h!]
\centering
\caption{Accuracy is measured by ``AUC'' on edge prediction tasks. P-values are calculated between AutoHEnsGNN$_{\text{Gradient}}$ and L-ensemble. AutoHEnsGNN$_{\text{Gradient}}$ significantly outperforms baseline methods on all datasets (p-values$<$0.05).}
\label{table:link_prediction}
\begin{tabular}{c|ccc}
\toprule
 Method               & Cora & Pubmed & Citeceer \\\midrule
Walkpooling~\cite{pan2021neural}     & 95.9 & 98.7   & 95.9    \\
S-VGAE~\cite{davidson2018hyperspherical}          & 94.1 & 96.0     & 94.7     \\
sGraphite-VAE~\cite{di2020mutual}   & 93.7 & 94.8   & 94.1     \\
Graph InfoClust~\cite{mavromatis2020graph} & 93.5 & 93.7   & 97.0       \\
BANE~\cite{yang2018binarized}            & 93.5 & -      & 95.6     \\
ARGE~\cite{pan2018adversarially}            & 92.4 & 96.8   & 91.9     \\\midrule
D-enesmble      &   95.2$\pm$1.0   &   98.0$\pm$0.0     &      95.5$\pm$0.5    \\
L-ensemble      &   95.9$\pm$0.9   &    98.6$\pm$0.0   &       96.4$\pm$0.3   \\
Goyal~\etal~\cite{goyal2019graph}           &   96.7$\pm$0.3   &   96.2$\pm$0.2     &    96.4$\pm$0.3      \\\midrule
AutoHEnsGNN$_{\text{Adaptive}}$     &   97.3$\pm$0.3   &    99.7$\pm$0.0    &    \textbf{97.6$\pm$0.3}      \\
AutoHEnsGNN$_{\text{Gradient}}$     &  \textbf{97.4$\pm$0.1}    &   \textbf{99.8$\pm$0.0}     &   97.5$\pm$0.2 \\\bottomrule 
\end{tabular}
\end{table}

\begin{table}[h!]
\centering
\caption{Accuracy on the graph classification task. P-values are calculated between AutoHEnsGNN$_{\text{Gradient}}$ and L-ensemble. AutoHEnsGNN$_{\text{Gradient}}$ significantly outperform baseline methods (p-values$<$0.05).}
\label{table:graph_prediction}
\begin{tabular}{c|c}
\toprule
Method            & PROTEINS \\\midrule
GIN~\cite{xu2018powerful}         & 76.2     \\
GraphSAGE~\cite{hamilton2017inductive}   & 73.0       \\
MEWISPool~\cite{nouranizadeh2021maximum}   & 80.7    \\
U2GNN~\cite{nguyen2019universal}       & 80.0    \\
sGIN~\cite{di2020mutual}        & 79.0    \\
hGANet~\cite{gao2019graph}      & 78.7    \\
HGP-SL~\cite{zhang2019hierarchical}      & 84.9     \\\midrule
D-enesmble  &     84.8$\pm$0.5     \\
L-ensemble  &     84.9$\pm$0.6     \\
Goyal~\etal~\cite{goyal2019graph}       &   84.8$\pm$0.5       \\\midrule
AutoHEnsGNN$_{\text{Adaptive}}$  &     85.4$\pm$0.3     \\
AutoHEnsGNN$_{\text{Gradient}}$ &     85.6$\pm$0.3   \\\bottomrule 
\end{tabular}
\vspace{-2mm}
\end{table}

\section{Conclusion}
In this paper, we propose AutoHEnsGNN for graph tasks, which automatically builds a hierarchical ensemble model for robust and accurate prediction. We propose proxy evaluation and efficient algorithms AutoHEnsGNN$_{\text{Gradient}}$ and AutoHEnsGNN$_{\text{Adaptive}}$, which can improve the efficiency and enable the automatic search. AutoHEnsGNN has won 1st place in the AutoGraph Challenge for KDD Cup 2020. The extensive experiments further demonstrate the effectiveness of AutoHEnsGNN. For future work, we will try to combine NAS, which is used to search novel architectures, with AutoHEnsGNN to achieve better performance.

\section{Acknowledgements}
Jin Xu and Jian Li are supported in part by the National Natural Science Foundation of China Grant 61822203, 61772297, 61632016, Turing AI Institute of Nanjing and Xi'an Institute for Interdisciplinary Information Core Technology. We also thank Kaitao Song for the helpful discussion and anonymous reviewers for the constructive comments.

\appendix
\section{Reproducibility}\label{sec:reproducibility}
\subsection{Training Configurations}
\subsubsection{Proxy evaluation}\label{repro:config:proxy_eval}
For public datasets Cora, Citeseer, and Pubmed, we use the exact training configurations and hyper-parameters as the original implements. For the anonymous dataset in AutoGraoh Challenge, we use Adam~\cite{kingma2014adam} optimizer with $\beta_1$=0.9, $\beta_2$=0.98, $\epsilon$=10$^{-9}$, train the model with a weight decay rate 5e-4, dropout rate \{0.5, 0.25, 0.1\} and learning rate \{5e-2, 3e-2, 1e-2, 7.5e-3, 5e-3, 3e-3, 1e-3, 5e-4\}. The learning rate is decayed with rate 0.9 every 3 epochs. We apply the early stop with patience 20 epochs and save the model with the best validation accuracy. We perform grid search for the key hyper-parameters including learning rate, dropout rate, and variants of models (e.g., GraphSAGE-mean, GraphSAGE-pool~\cite{hamilton2017inductive}). The optimal configurations can be automatically searched. 

\subsubsection{AutoHEnsGNN}\label{repro:config:autohensgnn}
For AutoHEnsGNN$_{\text{Gradient}}$, the architecture weights $\alpha$ and $\beta$ are updated at the end of every epoch as described in Algorithm 1. Each sub-model directly uses the hyper-parameters (dropout rate, etc.) in the proxy evaluation stage. The $K$ is set to 3 for self-ensemble. We use Adam optimizer with learning rate \{5e-2, 3e-2, 1e-2, 7.5e-3, 5e-3, 3e-3, 1e-3, 5e-4\} for the model weights and learning rate 3e-4 for architecture weights. The learning rate for the model weights is searched based on the validation accuracy. We apply the early stop with patience 5 epochs and obtain the $\alpha$ and $\beta$ when the model achieves the best validation accuracy. To reduce GPU memory consumption, we use proxy model with $M_{\text{proxy}}$=50\% for GPU memory reduction. For AutoHEnsGNN$_{\text{Adaptive}}$, $\alpha$ for each self-ensemble is decided by grid search and $\beta$ is decided by Equation~\ref{eqn:adative_weight} with $\epsilon=3$, $\gamma=8000$ and $\lambda=5$. 

\subsubsection{L-ensemble}\label{repro:config:L-ensemble} Rather than averaging the scores of different models for the ensemble, another method is to learn the ensemble weight on the validation set. To learn the weight, we split the $20\%$ dataset from the training set as the validation set. Then we fix the model weight of different models and apply gradient update on the learnable ensemble weight until convergence. We use Adam~\cite{kingma2014adam} optimizer with $\beta_1$=0.9, $\beta_2$=0.98, $\epsilon$=10$^{-9}$, learning rate \{1e-3, 1e-2, 1e-1\} and do not use weight decay.

\begin{table}[h!]
\centering
\caption{An example of an anonymous dataset in AutoGraph Challenge of KDD Cup 2020.}
\label{table:data_sample}
\scalebox{0.9}{
\begin{tabular}{cllcclclccll}
\toprule
\multicolumn{12}{c}{\textbf{Node File} }                                                                                                   \\\midrule
\multicolumn{4}{c}{Training nodes index (Int)}   & \multicolumn{8}{c}{1,2,3,4,5,$\cdots$}                                              \\
\multicolumn{4}{c}{Test node index (Int)}        & \multicolumn{8}{c}{128,129,130,$\cdots$}   \\\midrule\midrule                                  
\multicolumn{12}{c}{\textbf{Edge File} }                                                                                                   \\\midrule
\multicolumn{4}{c}{Source\_index}                   & \multicolumn{4}{c}{Destination\_index}      & \multicolumn{4}{c}{Edge weight}          \\\midrule
\multicolumn{4}{c}{Int}                          & \multicolumn{4}{c}{Int}             & \multicolumn{4}{c}{Float}                \\\midrule
\multicolumn{4}{c}{0}                            & \multicolumn{4}{c}{62}              & \multicolumn{4}{c}{1}                    \\
\multicolumn{4}{c}{0}                            & \multicolumn{4}{c}{40}              & \multicolumn{4}{c}{1}                    \\
\multicolumn{4}{c}{1}                            & \multicolumn{4}{c}{41}              & \multicolumn{4}{c}{2}                    \\
\multicolumn{12}{c}{$\cdots$}      \\ \midrule\midrule                                                                                                      
\multicolumn{12}{c}{\textbf{Feature File}   }                                                                                              \\\midrule
\multicolumn{3}{c}{Node Index}    & \multicolumn{3}{c}{F0}        & \multicolumn{3}{c}{F1}       & \multicolumn{3}{c}{$\cdots$}       \\\midrule
\multicolumn{3}{c}{Int}            & \multicolumn{3}{c}{Float}        & \multicolumn{3}{c}{Float}       & \multicolumn{3}{c}{Float}                          \\\midrule
\multicolumn{3}{c}{0}              & \multicolumn{3}{c}{0.477758}  & \multicolumn{3}{c}{0.053875} & \multicolumn{3}{c}{0.729954}  \\
\multicolumn{3}{c}{1}              & \multicolumn{3}{c}{0.3422409} & \multicolumn{3}{c}{0.669304} & \multicolumn{3}{c}{0.0873657} \\
\multicolumn{3}{c}{2}              & \multicolumn{3}{c}{0.825985}  & \multicolumn{3}{c}{0.442136} & \multicolumn{3}{c}{0.987225}  \\
\multicolumn{12}{c}{$\cdots$}  \\ \midrule\midrule                                                                                                   
\multicolumn{12}{c}{\textbf{Label File}    }                                                                                               \\\midrule
\multicolumn{6}{c}{Node Index}                                     & \multicolumn{6}{c}{Class}                                    \\\midrule
\multicolumn{6}{c}{0}                                              & \multicolumn{6}{c}{1}                                        \\
\multicolumn{6}{c}{1}                                              & \multicolumn{6}{c}{3}                                        \\
\multicolumn{6}{c}{2}                                              & \multicolumn{6}{c}{1}                                        \\
\multicolumn{12}{c}{$\cdots$}                                                                                                            \\ \midrule\midrule  
\multicolumn{12}{c}{\textbf{Metadata File}    }                                                                                              \\\midrule
\multicolumn{6}{c}{Time Budget (s)}                                & \multicolumn{6}{c}{n\_class}                                 \\\midrule
\multicolumn{6}{c}{500}                                            & \multicolumn{6}{c}{7}   \\\bottomrule
\end{tabular}}
\vspace{-3mm}
\end{table}

\subsection{AutoGraph Challenge for KDD Cup 2020}

The Automatic Graph Representation Learning challenge (AutoGraph), the first AutoML challenge applied to Graph-structured data, is the AutoML track challenge in KDD Cup 2020 provided by 4Paradigm, ChaLearn, Stanford and Google. The challenge website could be found online~\footnote{\url{https://www.automl.ai/competitions/3}}. In this challenge, participants need to design a computer program capable of providing solutions to graph representation learning problems without any human intervention. The tasks are also constrained by the time budget. This challenge focuses on the problem of graph node classification tasks to evaluate the quality of learned representations. AutoGraph Challenge provides 15 graph datasets: 5 public datasets, which can be downloaded, are provided to the participants; 5 feedback datasets are also provided to evaluate the public leaderboard scores of solutions; 5 final datasets are used to evaluate the quality of solutions in the final phase. The features of datasets are anonymous and numerical. There are more than one hundred teams that participated in this competition. 
Our team ``aister'' won 1st place in the final phase and our code is also publicly available.

\begin{figure}[h!]
\centering
\includegraphics[width=0.3\textwidth]{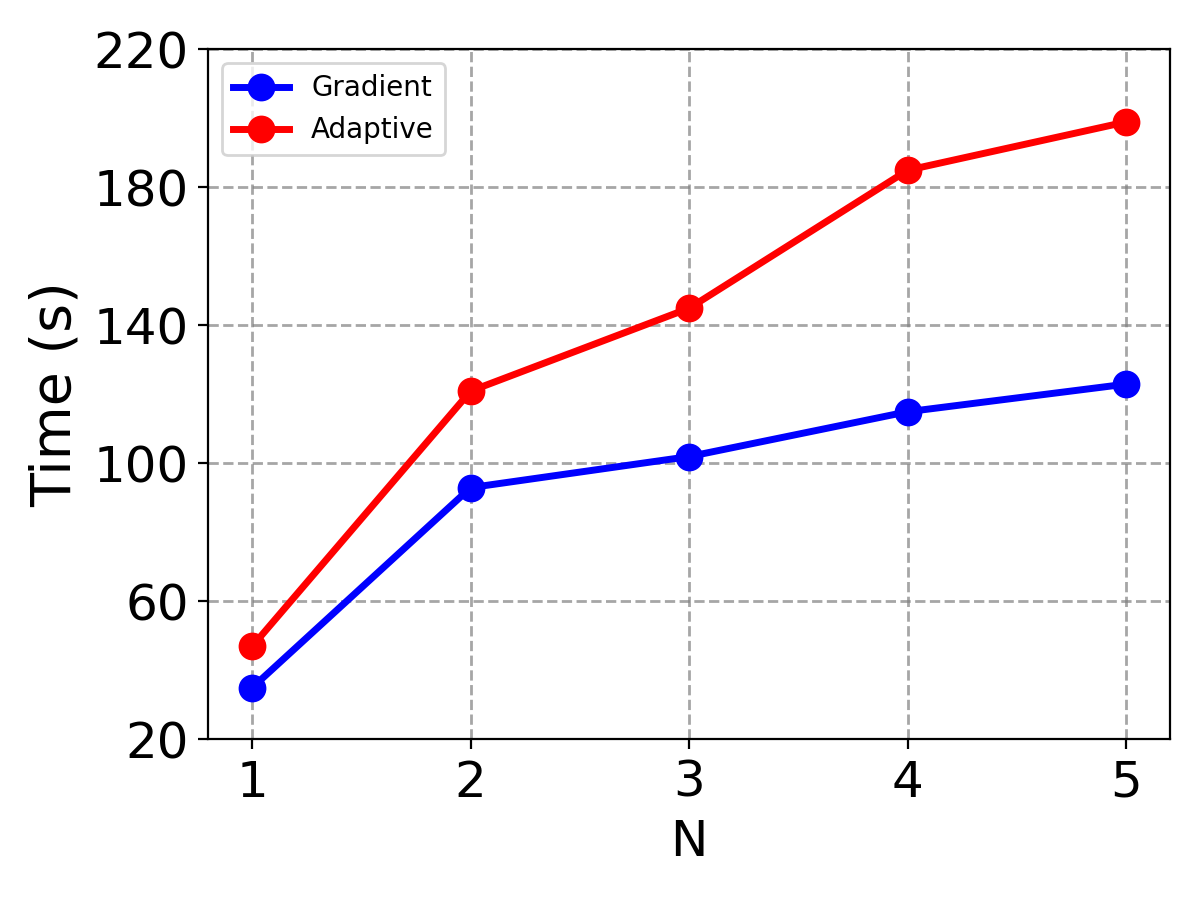}
\vspace{-4mm}
\caption{Time efficiency study by varying pool size $N$.}
\label{fig:n_two_time}
\vspace{-3mm}
\end{figure}

\subsection{Time efficiency study by varying the pool size $N$}
We study the time efficiency of AutoHEnsGNN on different pool sizes $N$ and conduct experiments on Cora datasets. We vary the pool size from 1 to 5 and measure the time consumption of the search stage on a single NVIDIA P40 GPU. As shown in Figure~\ref{fig:n_two_time}, the time consumption grows linearly for AutoHEnsGNN$_{\text{Adaptive}}$ as $N$ increases. However, due to the gradient optimization, AutoHEnsGNN$_{\text{Gradient}}$ does not significantly consume more time, which verifies the time efficiency of the bi-level optimization in Algorithm 1.

\clearpage
\bibliographystyle{IEEEtran}
\bibliography{AutoGNN_ref}
\end{document}